\documentclass[final]{cvpr}

\usepackage{times}
\usepackage{epsfig}
\usepackage{graphicx}
\usepackage{amsmath}
\usepackage{amssymb}
\usepackage{comment}
\usepackage{stfloats}

\newcommand{\lev}[1]{{\color[rgb]{1,0,0} {\Large #1}}}

\usepackage[pagebackref=true,breaklinks=true,colorlinks,bookmarks=false]{hyperref}

\def\cvprPaperID{****} 
\def\confYear{CVPR 2021}

\begin{document}

\title{\ Using a Supervised Method Without Supervision  for Foreground Segmentation}

\author{Levi Kassel\\
The Hebrew University of Jerusalem\\
Jerusalem, Israel \\
{\tt\small levi.kassel@mail.huji.ac.il}
\and
Michael Werman\\
The Hebrew University of Jerusalem\\
Jerusalem, Israel \\
{\tt\small michael.werman@mail.huji.ac.il}
}

\maketitle

\begin{abstract}
Neural networks are a powerful framework for foreground segmentation in video,  robustly segmenting moving objects from the background  in various challenging scenarios. The premier methods are those based on supervision requiring a final training stage on a database of tens to hundreds of manually segmented images from the specific static  camera.

In this work, we propose a method to automatically create a scene specific "artificial" database that is sufficient for training the supervised methods so that it performs better than current unsupervised methods. It is based on combining an unsupervised foreground segmenter to extract  objects from the training images and randomly inserting these  objects back in their correct location into a background image.

Test results are shown on  test sequences in CDnet.\footnote{This research was supported by the DFG.}
\end{abstract}

\section{Introduction}
Foreground segmentation/background subtraction is one of the central tasks in the field of computer vision due to its numerous  applications  including  surveillance of human activities in public spaces, traffic monitoring, and industrial machine vision.

Deep learning has become the principal  methodology in  foreground segmentation where each pixel in an image is classified as  foreground or background.

One of the great difficulties with supervised learning, especially in the field of foreground segmentation, is  the  labeling of data that usually requires extensive human hands-on work.

One of the known methods designed to eliminate the need to create data manually is to use synthetic data.
The problem with synthetic data in computer vision is the ability to create data that is similar enough to the real data so that the learned system can  be generalized to the original task with the real data.\\
This paper proposes substituting  the hand-labeled data needed for the premier (supervised) method with an automatically created "artificial" database. It is based on  an unsupervised weak foreground segmenter that extracts "good" objects from the training images and then to create the fine tuning data we randomly chose from these objects and insert them in their original position into a background image. It is especially pertinent to static cameras as the objects found are automatically in the "right" location, being of the "right" size, color and shape, and  in the "right" lighting conditions. In this way, we create an "scene specific" system in a unsupervised way.
This is applicable to a gigantic number of cameras, as cited in CNBC
\footnote{https://www.cnbc.com/2019/12/06/one-billion-surveillance-cameras-will-be-watching-globally-in-2021.html},
"One billion surveillance cameras will be watching around the world in 2021".

\section{Related work}

Deep learning methods are currently the premier methods in video foreground segmentation. 

Many  foreground segmentation methods work with one image at a time without taking into consideration temporal aspects between the frames. 
For example Lin et al. \cite{sakkos2018end} use the Fully Convolutional Semantic Networks (FCSN) \cite{long2015fully} for foreground segmentation.
The authors concatenate the current frame with the background image channel-wise and feed the 6-channel image to a FCN. Although their model achieves good results in simple videos, it fails in "noisy" environments and in detecting small objects.

In 2019, Lim and Keles presented FgSegNetv2\cite{lim2020learning}, 
a modified VGG 16 network is used as an encoder for  the  network,  obtaining  high-resolution  feature maps,  which  are  input  for  the  Feature Pooling Module (FPM) and consequently as input for the decoder, working with two Global Average Pooling (GAP) modules. This method used up to 200 annotated training frames of each scene  and achieves state of the art performance on the CDnet 2014 dataset\cite{wang2014cdnet}. Note that the performance of FgSegNet v2 drops dramatically when applied to unseen videos as shown by Tezcan et al. \cite{tezcan2020bsuv}.

Tezcan et al. \cite{9395443} presented the BSUV-Net 2.0 that mainly focused on data augmentation to enhance jitter, pan-tilt and intermittently-static objects especially trying to account for temporal variations that
naturally occur in video achieving state of the art results in the unsupervised case for various scenarios.

Synthetic data has  been employed  for data augmentation. For example, Richter et al.\cite{richter2016playing} and Ros et al. \cite{dosovitskiy2017carla} develop a virtual reality tool from the world of computer games for this purpose. A big advantage of this method is that you can produce endless samples of tagged data and you can create  situations as you wish. 

Dwibedi et al.\cite{dwibedi2017cut} used the power of synthetic data for instance  detection. They used a 'cut paste and learn' paradigm to cut objects and paste them on various background images to create a dataset for instance detection. They used Poisson blending and Gaussian smoothing on the pasted image to avoid  boundary artifacts. 

In order to benefit from  using  synthetic data it is necessary to use  foreground elements of the right categories; humans, vehicles, etc., with the the correct style of the elements of the foreground in a specific video.
And of course, the synthetic objects should be placed in the correct parts of the image with the right size and projection parameters. This entails a serious intervention for every scene and still there is a big difference between  the appearance  of synthetic data and reality.

These decisions and interventions are bypassed in our method because the foreground elements  are automatically extracted from the real images in the video sequence.
Therefore, our method is much more specific and appropriate to the environment in which the videos are shot.

\label{sec:format}

\section{Building the "supervised" training dataset}

The supervised foreground segmentation methods require a "supervised" labelled set of images.
Here we describe how we automatically generate a  video sequence dependent,  training set which is sufficient for the training of supervised based background subtraction architectures.   

\begin{figure}[ht]%
    \centering
    \includegraphics[width=8.87cm]{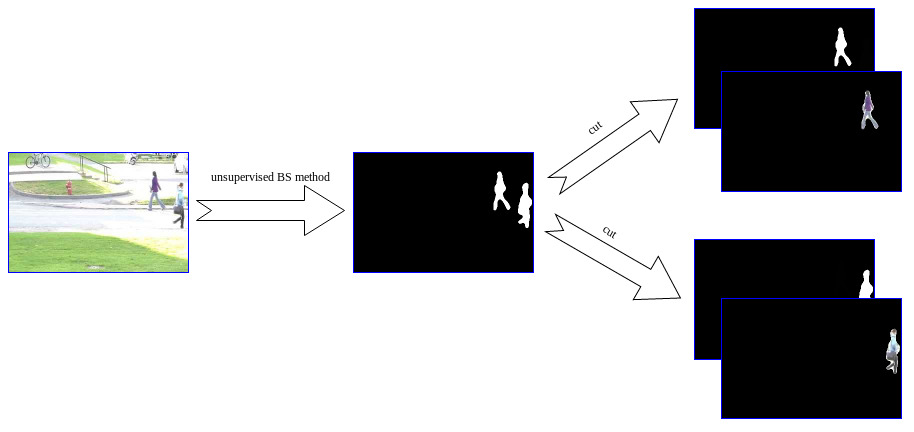}
    \includegraphics[width=8.87cm]{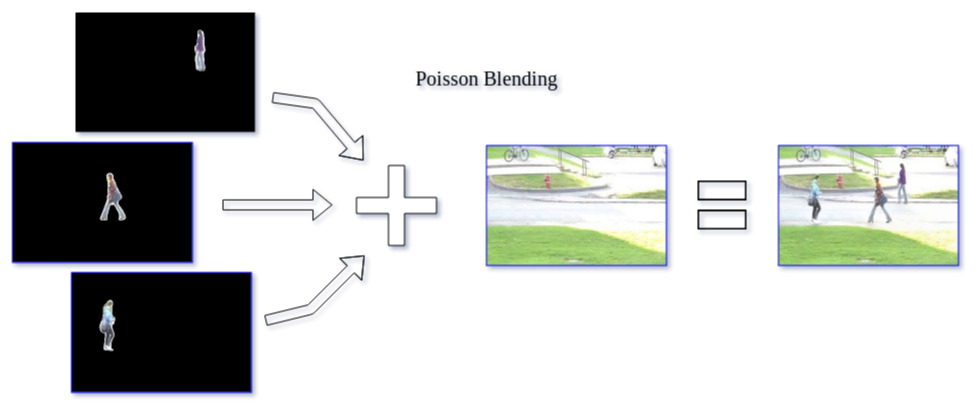}
    \caption{(top)  construction of the foreground object database. 
        (bottom) annotated image generation process 
    }%
    \label{fig:1}%
\end{figure}

\begin{table}
    \small
    \begin{tabular}{|l|c|c|c|c|c|c|}
    \hline
     \multicolumn{7}{|c|}{\textbf{BSUV-Net v2}} \\
    \hline
\textbf{Category}  & \textbf{25f} & \textbf{50f} & \textbf{100f} & \textbf{200f} & \textbf{500f} & \textbf{1000f}\\
\hline
\textbf{baseline}& 0.935& 0.955& 0.980& 0.984& \textbf{0.985}& 0.983 \\
\textbf{shadow}& 0.915& 0.939& 0.949& 0.960& \textbf{0.973}& 0.970 \\
\textbf{dynamic} & 0.900& 0.931& 0.952& \textbf{0.956}& 0.954& 0.950 \\
\hline
\end{tabular}
\caption{F-measure as a function of the number of images in augmented training set constructed from BSUV-Net v2 objects}
\label{table:BSUV-Net 2.0}
\end{table}

\begin{table*}[bp]
\centering
\begin{tabular}{|l|c|c|c|}
\hline\hline
\textbf{Category} & \textbf{baseline} & \textbf{shadow} & \textbf{dynamic} \\
\hline\hline
\hline
\multicolumn{4}{|c|}{\textbf{BSUV-Net v2}} \\
\hline
\textbf{Reference} & 0.962& 0.956& 0.905 \\
\hline
\multicolumn{4}{|c|}{\textbf{FgSegNet  V2}} \\
\hline
\textbf{BSUV } & 0.964& 0.950& 0.946 \\
\textbf{Objects from different scene \& random location} & 0.739& 0.673& 0.648 \\
\textbf{Same scene  objects \& random location}& 0.923& 0.893& 0.854 \\
\textbf{Same scene objects \& specific location} & {\bf 0.984}&  {\bf 0.960}& {\bf 0.956} \\
\hline
\end{tabular}
\caption{Different object selection and placement strategies, The BSUV results are followed by using FgSegNet V2 trained using   four different options,
the annotated frames from BSUV,
pasting  objects from other scenes in random locations, pasting  objects from BSUV on the same video  in random locations, and pasting  objects from BSUV on the same video  in their correct locations}
\label{table:random objects random places}
\end{table*}

We use the foreground segmentation results of an unsupervised method to extract objects from the video.
Subsets of these extracted objects are  randomly chosen and inserted into a background image to produce a training set.
As these objects are spatially positioned  exactly where they were found, cars are on the right roads, pedestrians on the traversed paths, and everything is where it really should be and with the correct projection and lighting,  without having  to analyze the scenario.

\begin{figure}%
    \centering
    \includegraphics[width=8.7cm]{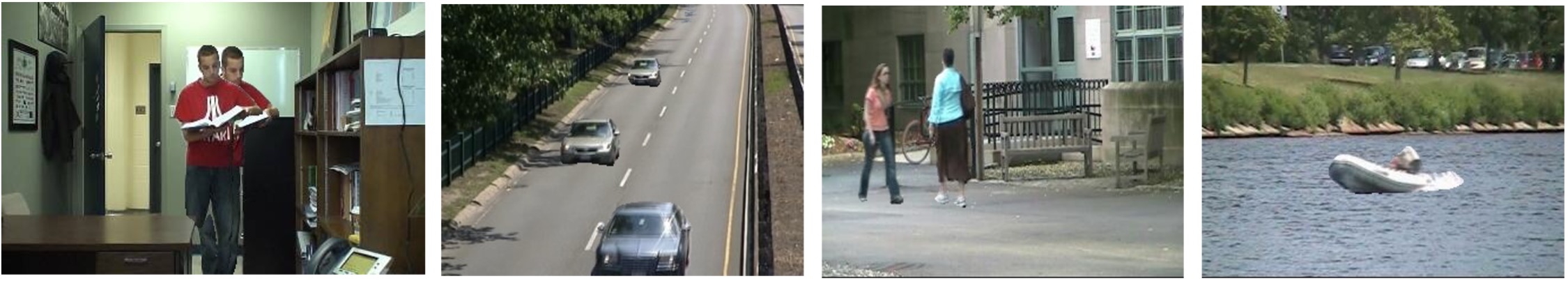}
    \includegraphics[width=8.7cm]{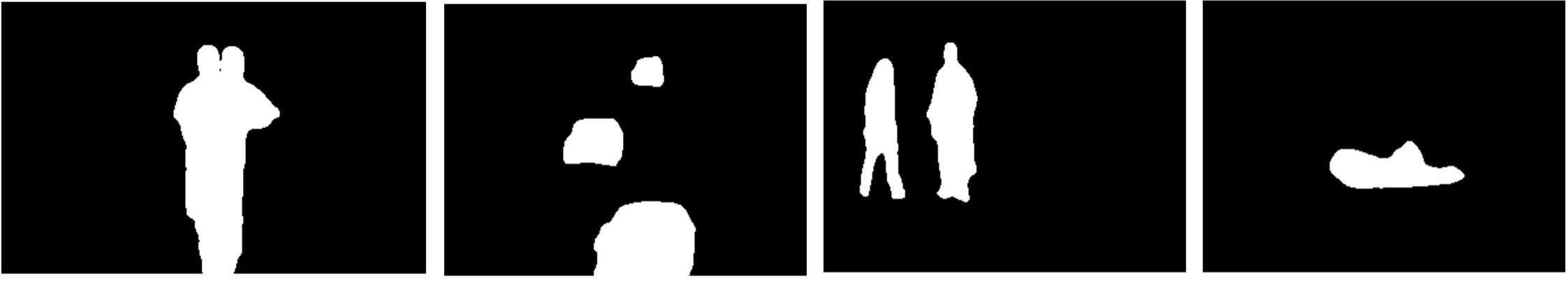}
    \caption{Examples from  our 'artificial'  database based on videos from the CDnet2014 dataset, (top)  generated images with the foreground objects, and (bottom)   corresponding foreground masks.}%
    \label{fig:2}%
\end{figure}

\begin{figure}%
    \centering
    \includegraphics[width=7cm]{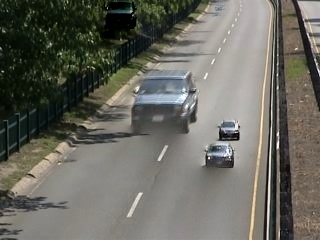}
    \includegraphics[width=7cm]{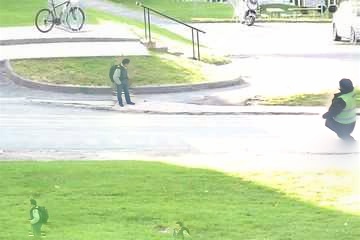}
    \caption{(top) randomly placed objects in 'highway', (bottom) objects of 'PETS2006' in 'pedestrian' scene }%
    \label{fig:3}%
\end{figure}

\subsection{Choosing the objects}
As we mentioned before, the foreground objects are extracted with unsupervised methods  in a 'cut \& paste' fashion such that a  foreground object and its mask are extracted to create a foreground database, see  Fig~\ref{fig:1}. In this work we use the state of the art unsupervised method on the CDnet dataset \cite{wang2014cdnet}, BSUV-Net v2 \cite{tezcan2021bsuv}.

\subsection{Building the dataset}
The construction of the database is  as follows:

Given the objects \begin{math} \mathcal{O} \end{math},
from the previous step and a background image \begin{math} \mathcal{I} \end{math},
randomly choose  objects from \begin{math} \mathcal{O} \end{math} and paste them in their original spatial locations onto \begin{math} \mathcal{I} \end{math}, see Fig~\ref{fig:1}. To avoid  boundary artifacts,
Poisson editing \cite{perez2003poisson}, as Dwibedi et al. suggested ,
was used. It turned out that  a Gaussian smoothing \cite{dwibedi2017cut} step was counterproductive. 
These images comprise the  dataset required for the supervised method,  examples are in Fig ~\ref{fig:2}. 

\section{Experiments}

We tested our method with  FgSegNet V2  by Lim et al. \cite{lim2020learning}, which is currently the top-ranked method  in the Change Detection 2014 Challenge\cite{wang2014cdnet}, SBI2015\cite{maddalena2015towards}, and UCSD Background Subtraction\cite{UCSD}.
FgSegNet V2 requires a final training stage using 25-200 labeled images from the video being evaluated and is thus a supervised method.
Our contribution is to use automatic object detection on the  images used for fine-tuning converting  the method to being  completely unsupervised but still scenario dependant.

Our method is especially relevant for static cameras so we tested  three of the datasets from CDnet: baseline, shadow, and dynamic background. 
The background images are pixel-wise medians of a sequence of 50  frames.
The frames used to extract objects were the same 200 used by FgSegNet V2\cite{lim2020learning}.

The average F-measure dependence on the number of augmented frames of the training set is shown in Table \ref{table:BSUV-Net 2.0}. Surprisingly more images is not always better, maybe due to inserting too many false positives.

To understand the power of our copy and correct location paste method in a static camera environment, we carried out two more experiments. In one we used the objects that were taken from the same scene but placed in random positions and thus do not take into account the natural location where they should be and in the other  we pasted objects from other scenes, see Fig  \ref{fig:3} for an illustration.
In both cases there was a decrease in the quality of the results, see Table \ref{table:random objects random places}.

To see that our method works with  different unsupervised methods we carried out the same  test based on the semanticBGS\cite{braham2017semantic} algorithm and saw similar improvements, see Table \ref{table:SemanticBGS}. 

\begin{table}[h]
    \small
    \begin{tabular}{|l|c|c|c|c|c|c|}
    \hline
    \multicolumn{7}{|c|}{\textbf{SemanticBGS}} \\
    \hline
\textbf{Category}  & \textbf{25f} & \textbf{50f} & \textbf{100f} & \textbf{200f} & \textbf{500f} & \textbf{1000f}\\
\hline

\hline
\textbf{baseline}& 0.943& 0.953& 0.976& 0.985& \textbf{0.985}& 0.980 \\
\textbf{shadow}& 0.917& 0.943& 0.959& 0.963& \textbf{0.970}& 0.968 \\
\textbf{dynamic} & 0.895& 0.933& 0.952& \textbf{0.953}& 0.942& 0.906 \\
\hline
\end{tabular}
\caption{F-measure as a function of the number of images in augmented training set constructed from SemanticBGS objects}
\label{table:SemanticBGS}
\end{table}

\vspace{-7mm}
\section{Summary}
We showed how a supervised foreground segmentation algorithm can be made unsupervised by replacing the supervision with augmented frames based on  objects  extracted with an unsupervised algorithm. This leads to a state of the art unsupervised foreground segmentation.

\newpage

{\small
\bibliographystyle{ieee_fullname}
\bibliography{main}
}

\end{document}